\documentclass{article}
\usepackage{arxiv}

\usepackage{graphicx}
\usepackage{subfig}
\usepackage{booktabs}
\usepackage{multirow}
\usepackage{url}
\usepackage{array}
\usepackage{amsfonts}
\usepackage{amsmath}
\usepackage{amssymb}  
\usepackage{bm} 

\usepackage[section]{placeins}

\AtBeginDocument{%
  \providecommand\BibTeX{{%
    \normalfont B\kern-0.5em{\scshape i\kern-0.25em b}\kern-0.8em\TeX}}}

\title{Correction of Faulty Background Knowledge based on Condition Aware and Revise Transformer for Question Answering}

\author{
    Xinyan Zhao \\
  University of Science and Technique of China\\
  Hefei, Anhui 230027\\
  \texttt{sa516458@mail.ustc.edu.cn} \\
   \And
 Xiao Feng \\
  University of Science and Technique of China\\
  Hefei, Anhui 230027\\
  \texttt{fx1995@mail.ustc.edu.cn} \\
  \And
 Haoming Zhong \\
  WeBank.com\\
  Shenzhen, Guangdong China\\
  \texttt{hmzhong@webank.com} \\
  \And
  Jun Yao \\
  WeBank.com\\
 Shenzhen, Guangdong China\\
  \texttt{junyao@webank.com} \\
  \And
 Huanhuan Chen \\
  University of Science and Technique of China\\
  Hefei, Anhui 230027\\
  \texttt{hchen@ustc.edu.cn} \\
}

\begin{document}
\maketitle

\begin{abstract}
The study of question answering has received increasing attention in recent years. This work focuses on providing an answer that compatible with both user intent and conditioning information corresponding to the question, such as delivery status and stock information in e-commerce. However, these conditions may be wrong or incomplete in real-world applications. Although existing question answering systems have considered the external information, such as categorical attributes and triples in knowledge base, they all assume that the external information is correct and complete. To alleviate the effect of defective condition values, this paper proposes condition aware and revise Transformer (CAR-Transformer). CAR-Transformer (1) revises each condition value based on the whole conversation and original conditions values, and (2) it encodes the revised conditions and utilizes the conditions embedding to select an answer. Experimental results on a real-world customer service dataset demonstrate that the CAR-Transformer can still select an appropriate reply when conditions corresponding to the question exist wrong or missing values, and substantially outperforms baseline models on automatic and human evaluations. The proposed CAR-Transformer can be extended to other NLP tasks which need to consider conditioning information.
\end{abstract}


\section{Introduction}

Question answering (QA), which intends to provide concise, direct answers to user queries based on abundant external information, has recently become a major focus of natural language processing (NLP) research. According to the type of external information, QA tasks can be roughly divided into two groups: text-QA and knowledge based QA (KBQA). The former utilizes information that in form of text such as web documents and the latter utilizes the knowledge from structured knowledge bases (KB) such as freebase \cite{bollacker2008freebase}, sales and marketing databases \cite{gao2019neural} and so on. Recently, neural approaches are exploited to represent questions and external information using continuous vectors so that the inference can be performed in a hidden semantic space \cite{bordes2014question,yang2014embedding,dong2015question,yin2016simple,yu2017improved,wang2018apva,seo2016bidirectional,liu2018stochastic}.  Neural approaches have advantages in semantic understanding since various expressions with similar semantic meaning have similar continuous representations in the hidden semantic space.

Although previous studies have made substantial progress, a major drawback of previous QA agents is that they assume the information from text or KBs is correct and complete, and thus they focus on how to utilize the information better. However, it might be unreasonable to suppose that the information is perfect in real-world applications. For example, many KBs are constructed by automatic extraction and might contain some mistakes due to lack of manual correction. In personalized dialog system, the user profile attributes may be incomplete due to the privacy policy. In addition, information delay, inaccuracy and missing in logistics, inventory and transaction are common in e-commerce applications.
\begin{table*}[!htbp]
\renewcommand{\arraystretch}{1.3}
\caption{Questions with different conditions responded by CAR-Transformer.}
\label{Examples}
\centering
\begin{tabular}{p{15em}p{14em}p{15em}}
\toprule
\bfseries Question & \bfseries Conditions & \bfseries Answer \\
\midrule
I need this item by Friday morning. Please could you send it ASAP? & Unshipped, expedite
delivery & We are now marked your order as
URGENT ORDER in our system
and we will try our best to delivery
it to you on time.\\ 
\cline{2-3}
 & Shipped, normal delivery & We will try our best but we are not able to guarantee.\\
\hline
Thank you for the update, yes I will wait for the new stock to come in. & Instock & Dear Customer, many thanks for your reply. We shall ship your order as soon as the new stock arrives.\\
\hline
Shirt arrived looks good, but too big as have another shirt similar. I need 17" collar, so XL would be better. Not opened package, so can I swap it for a smaller one? & Shipped: Unknow, delivery status: unknown & Dear customer, to receive the order in time, we suggest you to return the existing order for refund and place a new order for the right size so that you don't have to wait for the return to be processed. please request your return from here: \textless link\textgreater. We shall process your return request within two working days.\\
\bottomrule
\end{tabular}
\end{table*}

This work attempts to alleviate the impact of defective conditions corresponding to the question and provide a valid answer that is compatible with both user intent and the conditions. The conditions can be delivery status, delivery service type, stock information in e-commerce or user profile attributes in customer service. To be clear, Table \ref{Examples} shows several representative examples of the proposed model for customer service in e-commerce. The conditions in this application is order states. The first question in Table \ref{Examples} has different answers corresponding to different conditions. The second question has wrong condition value ``Instock'' and the third question has unknown condition values. But our model still gives appropriate responses. In general, there are two main challenges in our task. 
\begin{enumerate}
\item \textit{How to incorporate conditions into an end-to-end model.} Typical KBQA systems leverage knowledge in an explicit way, the answer is what retrieve from the knowledge base. However, in this study, conditions is to help answer selection instead of giving to the user directly.
\item \textit{How to deal with missing or wrong conditions.} Although existing defective condition values, it is not worthwhile to abandon the conditioning information completely. The model is expected to not only utilize the conditions but also reduce the effect of inferior condition values. Note that we don't know which condition of which sample is wrong in advance. 
\end{enumerate}

To tackle these challenges, this paper proposes condition aware and revise Transformer (CAR-Transformer). The overview of CAR-Transformer is shown in Figure \ref{framework}. Transformer \cite{vaswani2017attention} is an encoder-decoder framework that adopts self-attention mechanism to encode text instead of RNN or CNN structure. Since self-attention mechanism performs better in long distance dependence, Transformer is chosen as our baseline model. CAR-Transformer consists of four parts: conditions encoder, conditions reviser, dialogue encoder and classifier. We model the problem of conditions inference as a sequence generation problem and modify the Transformer architecture to dialogue encoder and conditions reviser. The dialogue encoder transforms conversation history and question into hidden representations. Then the conditions reviser generates revised conditions based on all original condition values and the dialogue representations. After that, each revised condition value is discrete and represented as a one-hot vector. Then the one-hot vectors are fed into the conditions encoder to get conditions embedding. Finally, the concatenation of conditions embedding and dialogue embedding is fed into the classifier to make a prediction over candidate responses. To capture the sophisticated interactions between features, the proposed conditions encoder and classifier adopt multi-layer neural network to take advantage of strong representation and generalization ability of deep learning. Experiments on a real-world dataset in customer service for e-commerce show that the CAR-Transformer can revise missing or wrong condition values to a certain degree and choose answer that accord with conditions. Automatic and human evaluations demonstrate CAR-Transformer achieves significant improvement as compared to all baseline methods. We also conduct experiments on personalized bAbI dataset to further verify the effectiveness of the CAR-Transformer. In summary, this paper makes the following contributions.
\begin{enumerate}
\item It proposes CAR-Transformer, a highly effective Transformer based conversational QA system. By revising and integrating condition values of question, CAR-Transformer outperforms several strong baselines.
\item By proposing the conditions reviser, CAR-Transformer is capable of revising wrong or incomplete condition values of question, which is more robust and practical in real-word applications compared to other QA systems. To our knowledge, this work is the first to discuss the treatment of defective external information in QA research. 
\item This paper explores how to represent and integrate the categorical information into the Transformer framework, therefore the prediction is compatible with both categorical attributes and the context. 
\end{enumerate}

The remainder of this paper is organized as follows. A review of related work is provided in Section II. Section III formulates the problem to be solved and presents CAR-Transformer. Section IV contains the introduction of the dataset used in this paper and abundant experiments. Finally, the conclusion and prospects for future work are provided in Section V.

\section{Related Work}

This section reviews the related work on sequence-to-sequence models and personalized dialog systems, which inspire the CAR-Transformer.

\subsection{Sequence-to-Sequence Models}
The sequence-to-sequence (Seq2Seq) model \cite{cho2014learning} is first proposed for machine translation task and also widely used in sequence generation task. Given a source $(x_1, x_2,...,x_T)$ and a target sequence $(y_1,y_2,..,y_{T^{\prime}})$, the model maximizes the conditional probability: $p(y_1 , ..., y_{T^{\prime}}|x_1 , ..., x_T)$. Seq2Seq model is in an encoder-decoder structure. The encoder summarizes a fixed-size vector representation from a variable-length input sentence, and the decoder generates sequences one by one based on the representation from the encoder and its previous outputs. The encoder and decoder can be specialized by RNN \cite{cho2014learning}, \cite{sutskever2014sequence}, \cite{bahdanau2014neural}, CNN \cite{gehring2017convolutional}, Transformer \cite{vaswani2017attention} and so on. RNN is able to process the temporal sequence. The advanced temporal analysis models include learning in the model space \cite{ChenTRY14,chen2014cognitive,chen2013model} and its variants \cite{gong2016model,chen2015model,gong2018sequential,gong2018multiobjective}. However, unlike RNNs, Transformers do not require that the sequence be processed in order. Particularly, Transformer only employs self-attention mechanism which computes a weighted sum by utilizing dot-products between elements of the input sequence \cite{lin2017structured}, \cite{bahdanau2014neural}, \cite{parikh2016decomposable}, \cite{kim2017structured} and achieves the state-of-the-art results on many natural language generation tasks. The difference between CAR-Transformer and prior works lies in that there is no temporal relationship between the order condition values which is the target sequence in this study. Therefore, the condition reviser utilizes all original conditions when revising one condition value and the revised condition values can be generated in parallel. While most of the Seq2Seq-based models generate words one by one by consuming the previously generated words as input.

\subsection{Attributes-aware Dialog Systems}
Learning the inherent attributes of dialogues explicitly is a way to improve the diversity and effectiveness of dialogue systems. Topic and personality are widely studied among different attributes. Xing \textit{et al.} \cite{xing2017topic} use Twitter LDA model to get the topic of dialog and then feed topic and input representations into a joint attention module to generate a topic-related response. Choudhary \textit{et al.} \cite{choudhary2017domain} divide each utterance in the dialogue into different domains and generates the domain and content of the next utterance accordingly. However, these studies require an extra component to infer the topic and suffer from the error in inferring topic. For personalized dialog systems, Li \textit{et al.} \cite{li2016persona} and Herzig \textit{et al.} \cite{herzig2017neural} project personal information into embedding space and then input the embedding vector into the decoder. Qian \textit{et al.} \cite{qian2017assigning} use an extra component to learn when employing the user profile. Yang \textit{et al.} \cite{yang2017personalized} and Zhang \textit{et al.}\cite{zhang2019neural} attempt to introduce personalized information to dialogs by transfer learning. These methods are all LSTM-based, which have inferiority when dealing with long sequences. For better capturing long-term dependency, the proposed model is Transformer-based. Besides, Joshi \textit{et al.}\cite{joshi2017personalization} and Luo \textit{et al.} \cite{luo2019learning} use an end-to-end memory network to select response in candidate sets. It is worth noting that all of the foregoing methods don't take into account whether flaws exist in personal information.

\section{The Proposed Method}
This section first formulates the investigated problem in this paper. Then Section \ref{Transformer layer} introduces the structure of transformer layer. Finally, a detailed description of CAR-Transformer is provided in Section \ref{sec-SAR-Transformer}.
\subsection{Problem Formulation}
Formally, let $(D,conditions\rightarrow r)$ denotes each training sample, where $D=\{u_1,...,u_n\}$ denotes a conversation with $n$ utterances. The last utterance is the question needing to be responded. $condition=\{c_1,...,c_m\}$ denotes $m$ condition values and $r$ denotes a candidate response. The goal of this work is to learn a classification model $G(D,conditions)$ to select an answer from candidate set for dialogue-conditions pair $(D,conditions)$ by computing the probability distribution of candidate answers.
\subsection{Transformer layer} \label{Transformer layer}
Transformer \cite{vaswani2017attention} is composed of a stack of identical layers, called transformer layers. Each transformer layer consists of a self-attention sub-layer followed by a feedforward sub-layer. Each sub-layer employs residual connection \cite{he2016deep} followed by layer normalization \cite{ba2016layer}. Let $X$ denotes the input of transformer layer, where $X$ is a sequence of $L$ $d$-dimensional vectors. The self-attention sub-layer first transforms $X$ into queries $Q=XW_Q$, keys $K = XW_K$ and values $V = XW_V$, where $W_Q$, $W_K$, $W_V$ are trainable $d \times d$ matrices. Each $L \times d$ query, key, and value matrix can be split into $HL \times d_h$ parts called attention heads, indexed by $h$, and with dimension $d_h = d/H$. This multi-heads attention mechanism allows the model to focus on different parts of the input sequence. Then the output of each head is calculated as:
\begin{equation}\label{...}
Attention(Q^h,K^h,V^h)=Softmax(\frac{Q^h,K^{h\top}}{\sqrt{d_h}})V^h.
\end{equation}
The outputs of each head are concatenated and linearly transformed to a $L$ by $d$ dimensional matrix by feedforward sub-layer:
\begin{equation}\label{...}
FFN(x)=max(0,xW_1+b_1)W_2+b_2.
\end{equation}
Since self-attention mechanism has no recurrent operation like RNN, Transformer keeps the absolute or relative position of a word by adding position encodings to the input embeddings of the bottom transformer layer. The position encoding is formated by:
\begin{equation}\label{...}
p(pos, k)=\left\{
\begin{array}{rcl}
sin(pos/10000^{k/d})     &      & \text{if k is even}\\
cos(pos/10000^{k/d})     &      & \text{if k is odd},\\
\end{array} \right.
\end{equation}
where $pos$ is the position of a word and $k$ is the index of the input dimension.
\subsection{CAR-Transformer} \label{sec-SAR-Transformer}
\begin{figure*}[!htbp]
\centering
\includegraphics[width=6in]{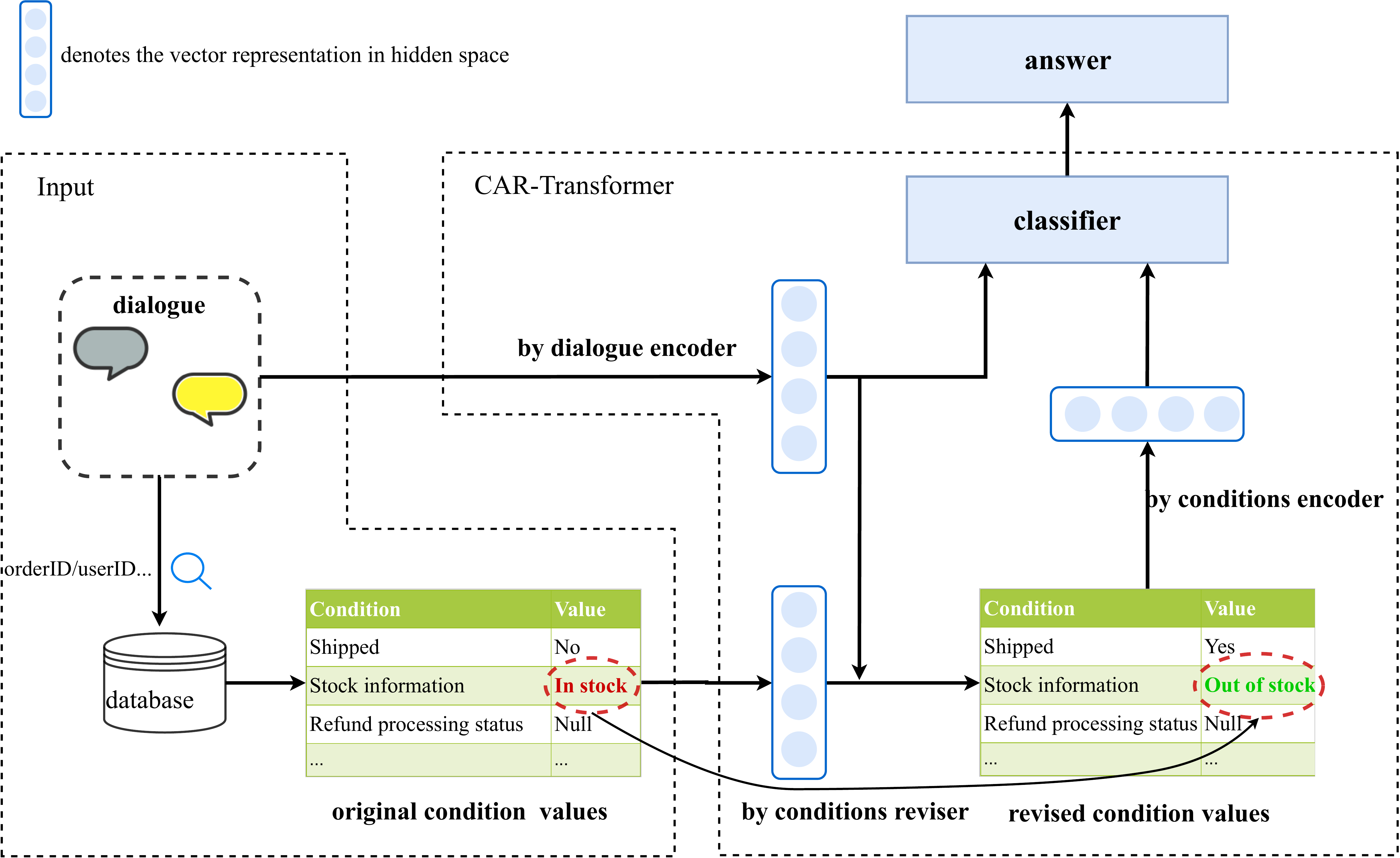}
\caption{Overview of the proposed CAR-Transformer. CAR-Transformer first revises conditions according to original condition values and dialogue. For example, the original order condition value "No" of "Return goods received" is wrong and the model revises it correctly. Then CAR-Transformer selects an answer that in line with the dialogue content and the conditions.}
\label{framework}
\end{figure*}
Overview of CAR-Transformer is shown in Figure \ref{framework}. CAR-Transformer first revises conditions of the question according to the original condition values and dialogue context. Then CAR-Transformer selects an answer in line with the dialogue content and the revised conditions. CAR-Transformer is consisted of four components:
\begin{itemize}
\item \textbf{conditions encoder} encodes the $m$ conditions into a vector representation;
\item \textbf{dialogue encoder} summarizes the whole conversation into a sequence of vector representations;
\item \textbf{conditions reviser} generates revised condition values based on the original condition values and the dialogue representations;
\item \textbf{classifier} selects a candidate response based on the revised conditions embedding and the dialogue representations. 
\end{itemize} 
Since CAR-Transformer has two outputs: revised condition values and predicted label of response, the overall objective function is defined as following:
\begin{equation}\label{loss}
\mathcal{L}=\eta\mathcal{L}_c+(1-\eta)\mathcal{L}_r,
\end{equation}
where $\mathcal{L}_c$ denotes the sum of cross entropy loss of each condition value, $\mathcal{L}_r$ denotes the cross entropy between the predictive distribution and the true candidate response label and $\eta$ is a scalar to balance the two terms. 

\begin{figure}[!htbp]
\centering
\includegraphics[width=2in]{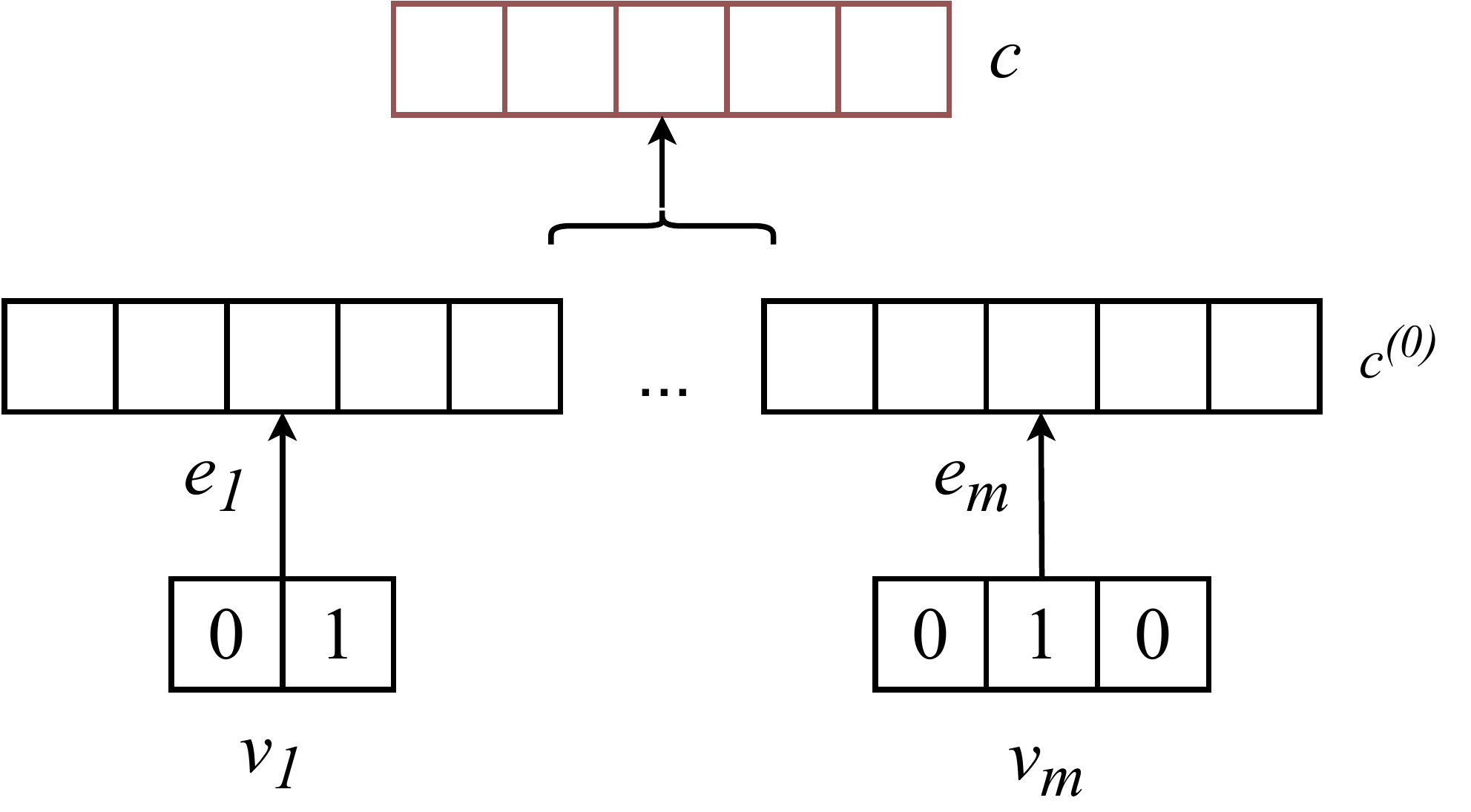}
\caption{The structure of conditions encoder.}
\label{stateencoder}
\end{figure}

\begin{figure}[!htbp]
\centering
\includegraphics[width=3in]{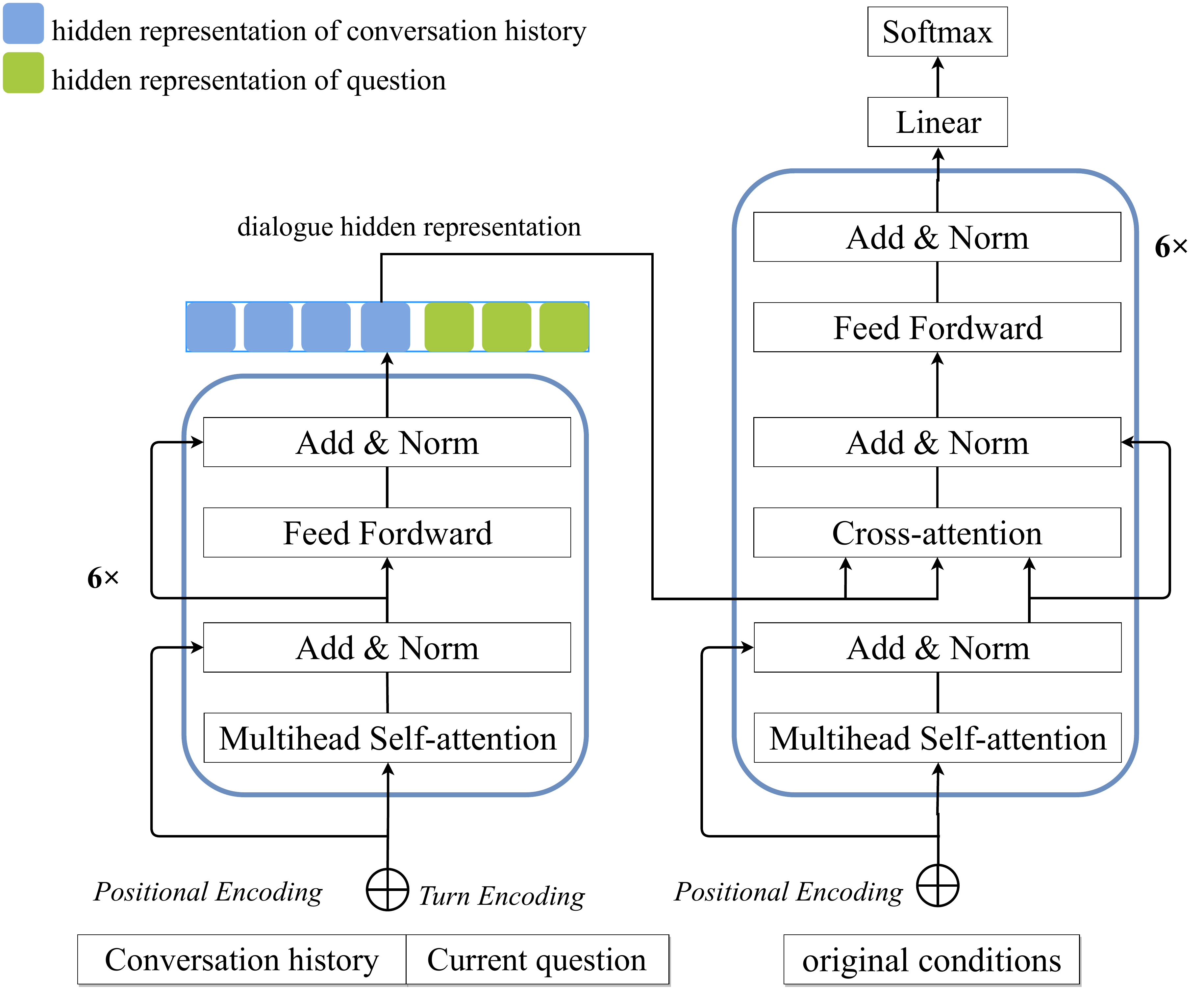}
\caption{Details of dialogue encoder (left part) and conditions reviser (right part).}
\label{statereviser}
\end{figure}

Figure \ref{stateencoder} highlights the structure of conditions encoder. Figure \ref{statereviser} illustrates the structure of dialogue encoder and conditions reviser. Details of each component will be described in the following subsections.
\subsubsection{Conditions Encoder}
Objective of this component is to represent the conditions as a vector and capture the interaction between conditions. Each condition value is first converted into one-hot vector $v_i$. Then each one-hot vector is compressed into embedding space. The embeddings are concatenated as the output of the embedding layer:
\begin{equation}\label{...}
c^{(0)}=[e_1,e_2,...,e_m],
\end{equation}
where $e_i$ is the embedding of $i$-th condition. Although the lengths of different one-hot condition vectors
can be different, the embeddings of different condition are of the same size $s$, so the size of $c^{(0)}$ is $m \times s$. Finally $c^{(0)}$ is fed into a fully connected layer:
\begin{equation}\label{...}
c=\sigma(W_{c}c^{(0)}+b_c),
\end{equation}
where $c$ is the final representation of $m$ conditions and $\sigma$ is an activation function. $W_c$ and $b_c$ is weight and bias of the fully connected layer, respectively. The size of $c$ is a hyperparameter that determines the dimension of the conditions embedding.
\subsubsection{Dialogue Encoder}
Dialogue $D$ in each sample is split into two parts: dialogue history $History=\{u_1, ...,u_{n-1}\}$ that contains the first $n \mathbf{-} 1$ utterances and the $n$-th utterance $u_n$. $u_n$ is the question needing to be responded. All tokens in $D$ are concatenated and an end-of-utterance delimiter is inserted between every two utterances. For each token $w$ in $D$, the input embedding is the sum of its word embedding, position embedding and turn embedding:
\begin{equation}\label{...}
I(w)=WE(w)+PE(w)+TE(w).
\end{equation}
Dialogue encoder follows the same way in normal Transformer \cite{vaswani2017attention} to compute the word embedding $WE(w)$ and position embedding $PE(w)$. Tokens from $History$ share the same turn embedding and tokens from $u_n$ share the same turn embedding. Then the input embeddings are forwarded into dialogue encoder to get the hidden representation $\mathbf{z}$, which has the same length and dimension as input. The dialogue encoder is composed of a stack of 6 transformer layers. Readers can refer to Section \ref{Transformer layer} for details of the transformer layer.
\subsubsection{Conditions Reviser}
This component generates condition values based on dialogue hidden representations (hidden representations of conversation history and current question) and original condition values. The condition reviser is also composed of a stack of 6 identical sub-layers. Each layer has a self-attention sub-layer, feedforward sub-layer and an extra cross-attention sub-layer between self-attention sub-layer and feedforward sub-layer. The structure of each sub-layer is now described in detail. In normal Transformer, the self-attention sub-layer in decoder is modified to masked self-attention sub-layer to prevent positions from attending to subsequent positions. Conversely, the self-attention sub-layer in condition reviser takes attention over all positions by removing the mask encoding. This is because there is no temporal relationship between condition values and a condition may be correlated with conditions. Inspired by the attention mechanism in sequence-to-sequence models such as\cite{wu2016google},\cite{bahdanau2014neural},\cite{wang2016chinese},\cite{gehring2017convolutional}, the cross-attention sub-layer employs multi-head attention over all dialogue hidden representations. The cross-attention sub-layer follows the same structure as normal self-attention sub-layer, the query vectors are transformed from the outputs of the previous self-attention sub-layer, but the key vectors and value vectors are transformed from dialogue hidden representations. Finally, the outputs of cross-attention sub-layer are fed to feedforward sub-layer to get the final hidden representation vectors of this sub-layer. At the top of condition reviser, linear transformation and softmax function are applied to convert each hidden representation into predicted value distribution corresponding to each condition. In general, one condition value is revised by referring to the whole dialogue and all original conditions: 
\begin{equation}\label{...}
c_i^\prime = f(\mathbf{z},c_1,...,c_m),
\end{equation}
where $\mathbf{z}$ denotes the dialogue hidden representations, $c_i^\prime$ is revised condition value of $i$-th condition and $f$ is a nonlinear function.

\subsubsection{Classifier}
The classifier is constructed as a Multilayer Perceptron (MLP). The input of the MLP is the concatenation of conditions embedding, hidden representation for the first token of conversation history and hidden representation for the first token of question. The classifier outputs the predictive distribution of candidate responses. In the future, more elegant probabilistic classifiers, such as probabilistic classification vector machine and its variants \cite{chen2009probabilistic,chen2013efficient,lyu2019multiclass,jiang2017scalable} would be employed to produce real probabilistic outputs. Another direction is to employ neural network ensemble algorithms \cite{chen2010multiobjective,Soares12} for possible better performance, though its probabilistic outputs could be achieved by incorporating with Bayesian methods \cite{chen2009predictive,chen2009regularized}.

\section{Experiment and analysis}
\subsection{Dataset Description}
\begin{table*}[!htbp]
\renewcommand{\arraystretch}{1.3}
\caption{Introduction of order conditions and its possible values}
\label{states}
\centering
\begin{tabular}{p{8em}p{15em}p{12em}}
\toprule
\bfseries Conditions & \bfseries Possible values & \bfseries Description \\
\midrule
Shipped & Yes/No & Does this order has been shipped? \\
\hline
Delivery status & Null/Normal/Delay/Deliver failed/Redelivery/Missing/Unknown & If the order has been shipped, the status of delivery. \\
\hline
Consignee's area & US/NG/GB/Other site & Area of the consignee.\\
\hline
Delivery service type & Null/Expedite service/Normal & Type of delivery service chosed by buyer.\\
\hline
Stock information &In stock/Out of stock &Are the items in the order still available?\\
\hline
Return goods received & Null/Yes/No & Dose seller receive the goods sent back from the buyer? \\
\hline
Refund processing status & Null/Unprocessed/Refunded & Whether pass buyer's refund application? \\
\bottomrule
\end{tabular}
\end{table*}

\begin{figure}[!htbp]
\centering
\includegraphics[width=3in]{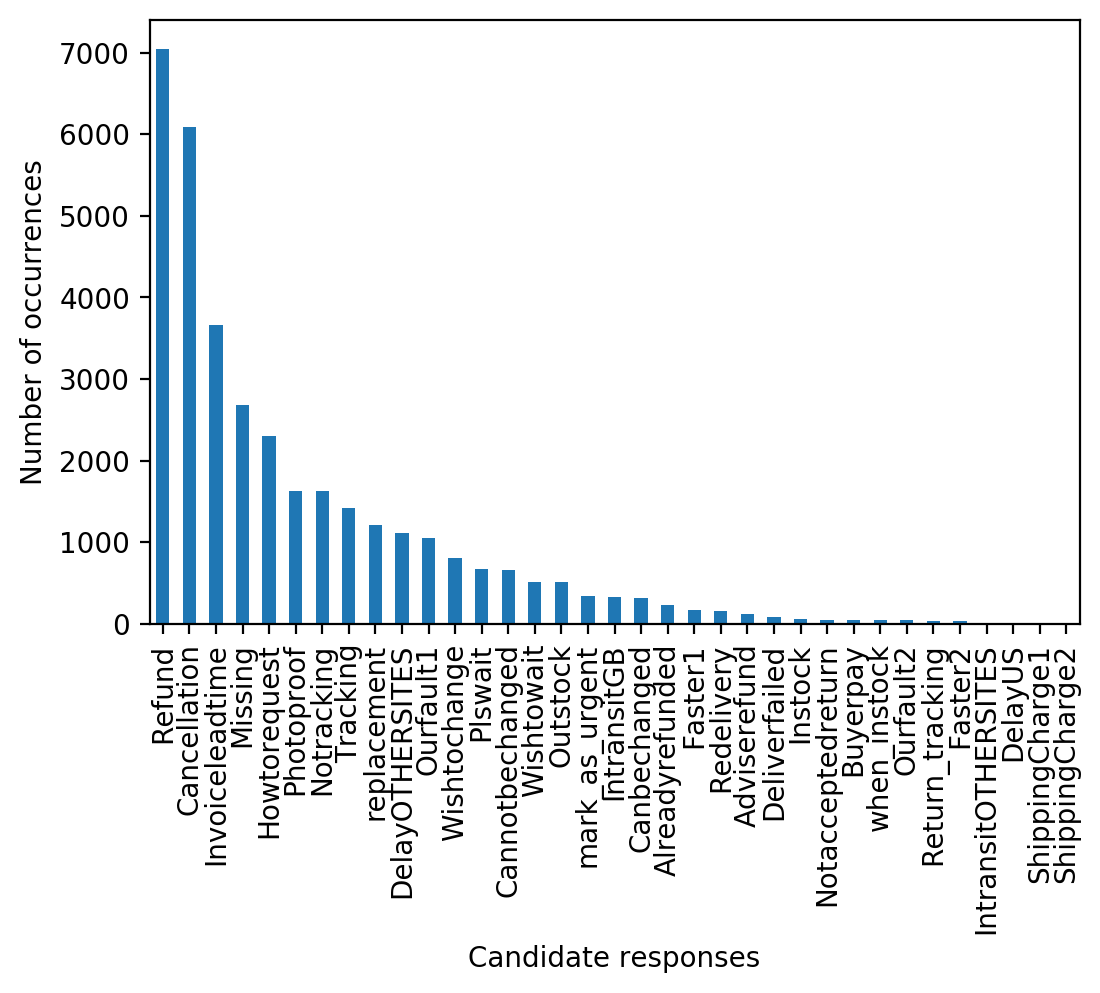}
\caption{Distribution of candidate responses.}
\label{distribution}
\end{figure}

Dialogs used in this section are collected from an online customer service system for e-commerce provided by our collaborator. The final dataset includes 35928 dialogs. The dataset is randomly divided into training (25150), validation (3593) and testing (7185) sets. Each dialog has seven order condition values associated with its order ID. The conditions and its possible values are shown in Table \ref{states}. ``Null'' means the absence of the value of this condition. For example, if an order has not been shipped, the order does not have delivery status, of course. Besides, ``Unknown'' means the value of this condition is missing due to some reasons. Every dialog has 35 candidate replies. Abbreviations are used to represent each candidate reply in the following parts. The distribution of candidate responses in this dataset is shown in Figure \ref{distribution}. The statistics of dataset is shown in Table \ref{Dataset statistics}, one can find that the utterance length in this dataset is relatively long.

\begin{table}[!htbp]
\renewcommand{\arraystretch}{1.3}
\caption{Dataset statistics}
\label{Dataset statistics}
\centering
\begin{tabular}{p{6em}p{5em}p{5em}p{5em}}
\toprule
\bfseries Dataset & \bfseries utterance average length & \bfseries Questions longer than 40 & \bfseries Questions with defective condition values \\
\midrule
Training set & 27.10 & 11.72\% & 19.63\% \\
Validation set & 27.35 & 11.82\% & 20.78\% \\
Testing set & 28.92 & 12.22\% & 20.53\% \\
\bottomrule
\end{tabular}
\end{table}

\subsection{Metrics}
\subsubsection{Automatic Evaluations}
Bilingual Evaluation Understudy (BLEU) \cite{papineni2002bleu} is widely used as an automatic evaluation of text generation systems. BLEU-1, BLEU-2, BLEU-3, and BLEU-4 scores are taken as the automatic evaluation metrics for the answers given by different methods:

\begin{equation}
BLEU{\text -n}=BP\cdot exp\left(\frac{1}{n}\sum_{i=1}^{n}logp_i\right),
\end{equation}
where $BP$ is the \textit{brevity  penalty  value}, which equals to 1 if the  total length of the resulting response is longer than that of the reference response or equals to the ratio between those two lengths. $p_i$ measures the overlapping between the bag of i-grams appearing in the resulting sentences  and that of i-grams appearing in the reference sentences. The value of BLEU score is between 0-1. The higher it is, the higher the precision of n-grams. For classification models, answer selection accuracy is also used as the evaluation metric.

\subsubsection{Human Evaluations}
Five customer service staffs were asked to label whether a reply is accord with order conditions and score satisfaction of the reply. Whether a reply is accord with order conditions is assessed by comparing the reply with ground truth condition values and two options (``Yes'' and ``No'') are optional. Reply satisfaction is an overall measurement, which reflects the response's fluency, availability and so on. There are five ratings for reply satisfaction. 300 samples are chosen randomly from the test set for human assessment, and all different categories of questions are guaranteed to appear in the set for human assessment.

\subsection{Baselines} 
\label{section:baselines}

CAR-Transformer is compared with the following methods. \textbf{Constant} replies ``Dear Customer, we have updated your order information to our fulfillment team.'' to all questions. \textbf{SC-LSTM} \cite{wen2015semantically} is a language generation model based on a semantically controlled Long Short-term Memory structure. By incorporating dialogue act one-hot vectors into the original LSTM \cite{gers2000learning} cell, SC-LSTM enables the generator to output the act-related text. Dialogue act one-hot vectors are replaced with order conditions one-hot vectors to adapt to the task of this paper. \textbf{TA-seq2seq}\cite{xing2017topic} utilizes topic information in chatbots by a joint attention mechanism. Topic words in \cite{xing2017topic} are replaced with our order conditions. \textbf{Split Memory Network}\cite{joshi2017personalization} is a modification of Memory Network \cite{weston2014memory} that enables personalization. Profile attributes and dialogue history are modeled in two separate memories. The profile attributes are replaced with order conditions. \textbf{BERT} \cite{devlin2018bert} is a Transformer-based pre-training model for language understanding. BERT summaries the input question as a fixed-dimensional pooled representation and this representation is transformed into label probabilities. \textbf{CA-Transformer} is a degraded version of CAR-Transformer without conditions revise component. 

The max length of input question is set to 50 and the max turn of history conversations is set to 2 for all models. The optimal setting of each model was selected by the BLEU scores on the validation set. We empirically set the hyperparameter $\eta$ in Equation \ref{loss} as 0.2. Both dimensions of word embedding and conditions embedding are 300 for all models.

\subsection{Results}
\begin{table*}
\caption{overall comparison results of different models}
\centering
\begin{tabular}{cccccccc}
	\toprule
    \multirow{2}{*}{Models} & \multicolumn{5}{c}{Automatic Evaluations} & \multicolumn{2}{c}{Human Evaluations} \\
	\cmidrule(r){2-6}  \cmidrule(r){7-8}
		& BLEU-1 & BLEU-2 & BLEU-3 & BLEU-4 & Accuracy & Rate of accord with conditions & Satisfaction \\
	\midrule
	constant   & 5.73\% & 3.70\% & 2.49\% & 2.10\% & 1.55\% & 58.08\%(0.26) & 1.38(0.36)\\
	SC-LSTM    & 37.97\% & 30.86\% & 27.64\% & 25.42\% & - & 39.93\%(0.21) & 1.26(0.20) \\
	TA-seq2seq & 58.95\% & 54.02\% & 51.64\% & 49.49\% & - & 71.33\%(0.10) & 3.26(0.19)\\
	Split Memory Network    & 52.55\% & 51.59\%& 50.88\% & 50.69\% & 50.60\% & 69.13\%(0.16) & 3.39(0.26) \\
	BERT & 67.70\% & 64.67\%& 62.64\% & 61.32\% & 59.29\% & 59.34\%(0.15) & 3.59(0.31) \\
	\midrule
	CA-Transformer & 75.23\% & 74.30\% & 73.45\% & 73.24\% & 73.15\% &75.27\%(0.11) & 3.76(0.07)\\
	CAR-Transformer & 86.53\% & 85.58\% & 84.88\% & 84.68\% & 85.60\% & 85.90\%(0.09) & 4.20(0.07)\\
	\bottomrule
\label{overall comparion results}
\end{tabular}
\end{table*}

The automatic evaluation and human evaluation results of different models are reported in Table \ref{overall comparion results}. For human evaluations, the mean and standard error of results given by five staff are reported. Based on the results, CAR-Transformer delivers the best performance of all the comparison methods on all metrics. 
Furthermore, the experimental results reveal the following observations.

\subsubsection{CAR-Transformer/CA-Transformer Can Better Leverages Conditions Information}

\begin{figure}[h]
\centering
\includegraphics[width=3.5in]{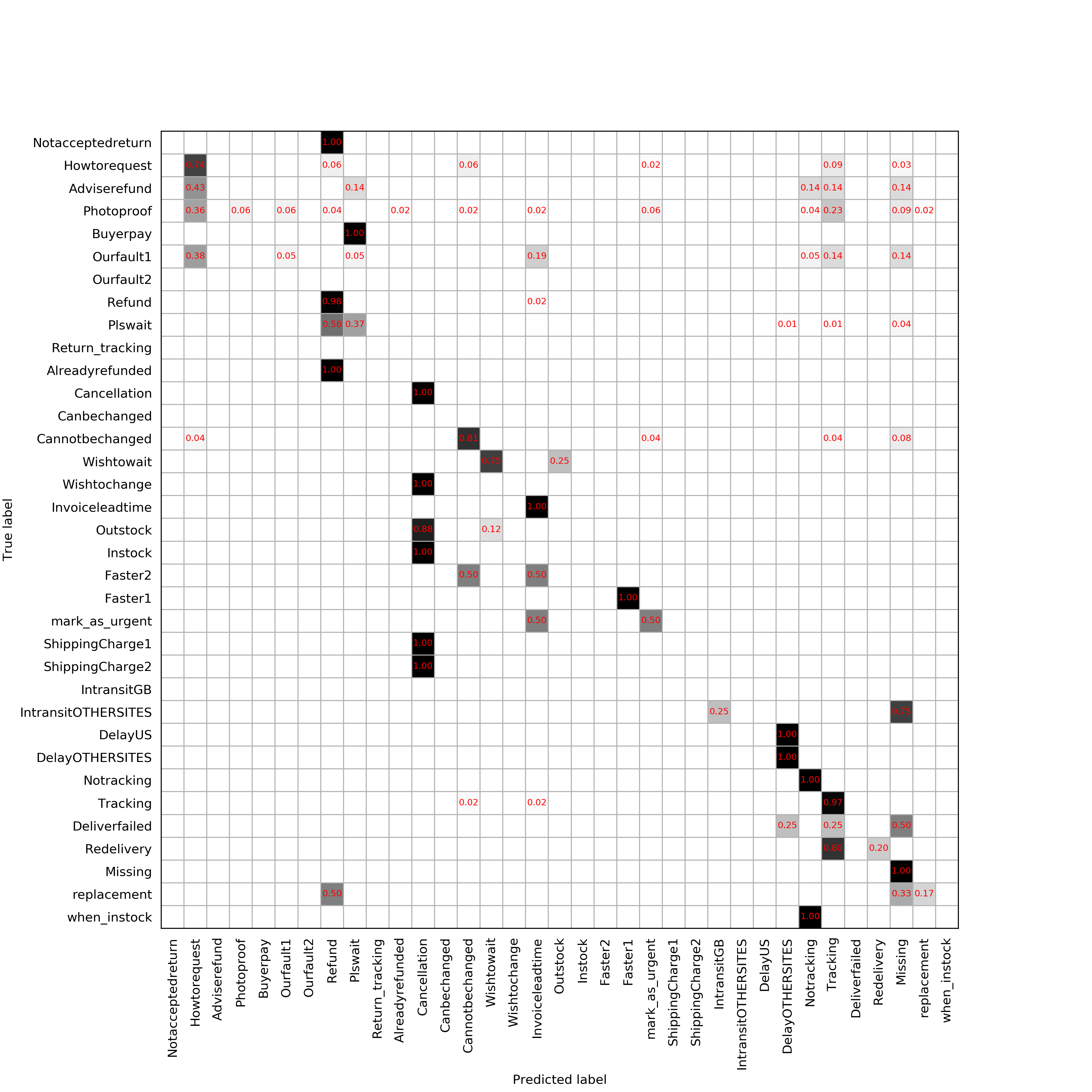}
\caption{Confusion matrix of CA-Transformer}
\label{SA-Transformer}
\end{figure}

\begin{figure}[h]
\centering
\includegraphics[width=3.5in]{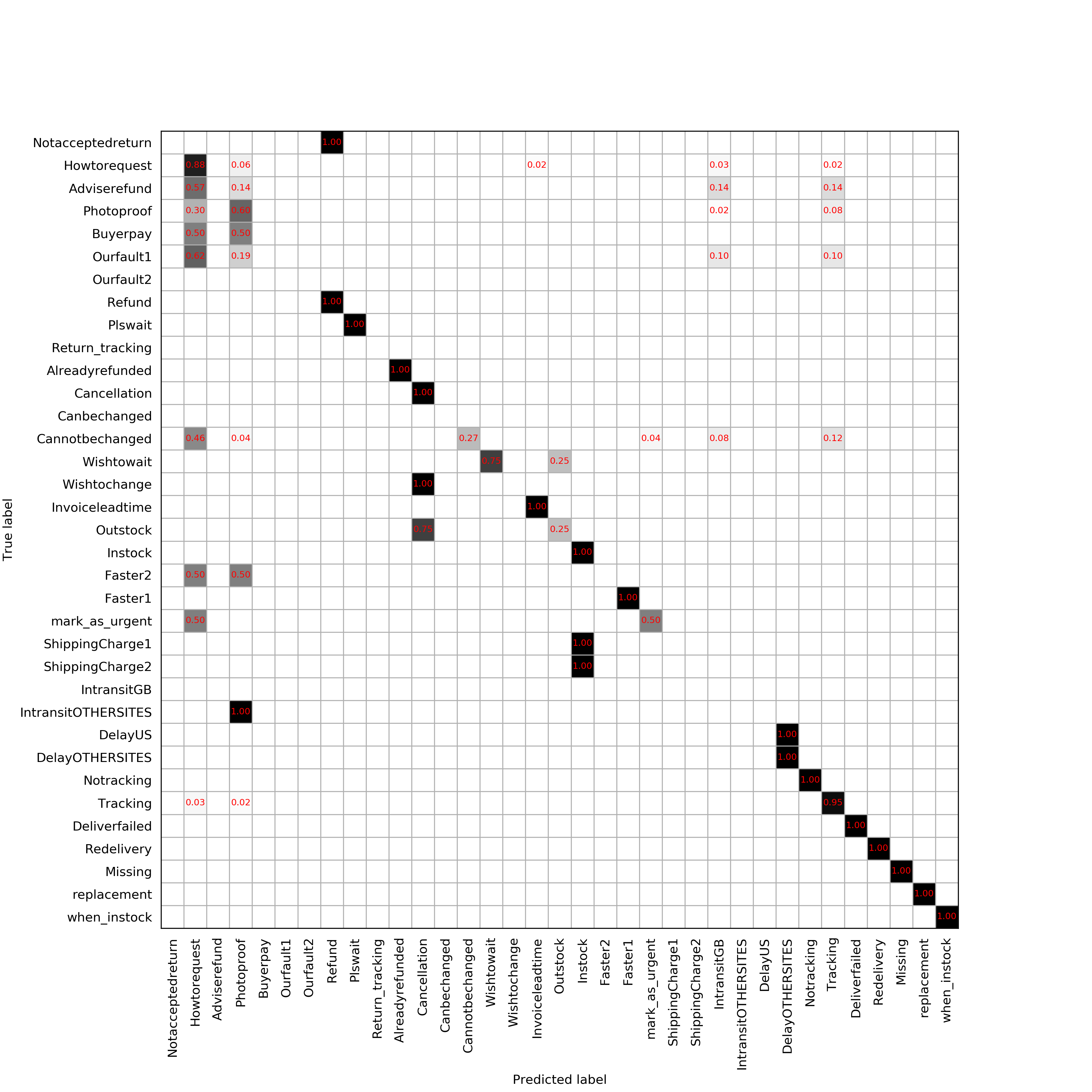}
\caption{Confusion matrix of CAR-Transformer}
\label{SAR-Transformer}
\end{figure}
\begin{figure}[h]
\centering
\includegraphics[width=3.5in]{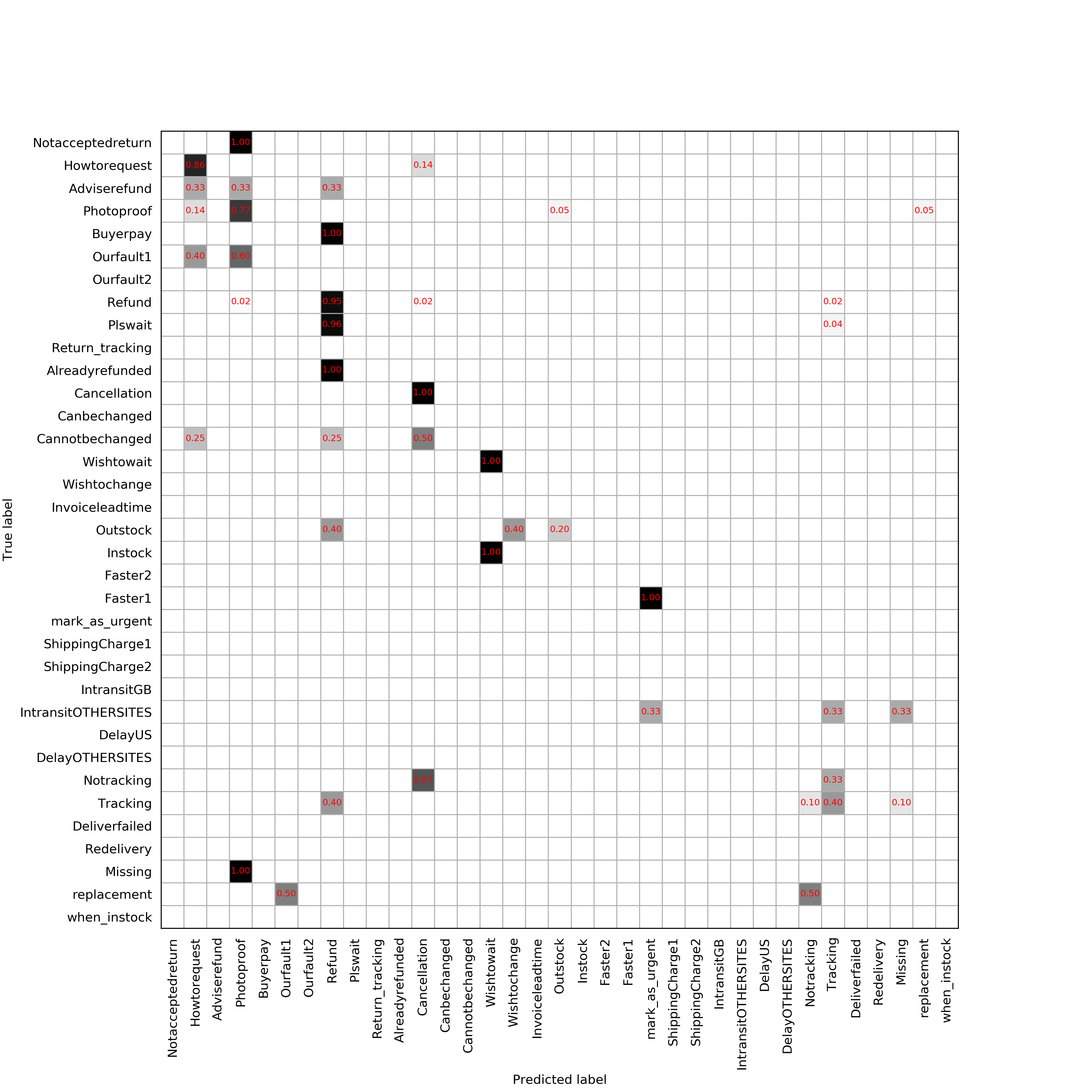}
\caption{Confusion matrix of BERT}
\label{bert}
\end{figure}
\begin{figure}[h]
\centering
\includegraphics[width=3.5in]{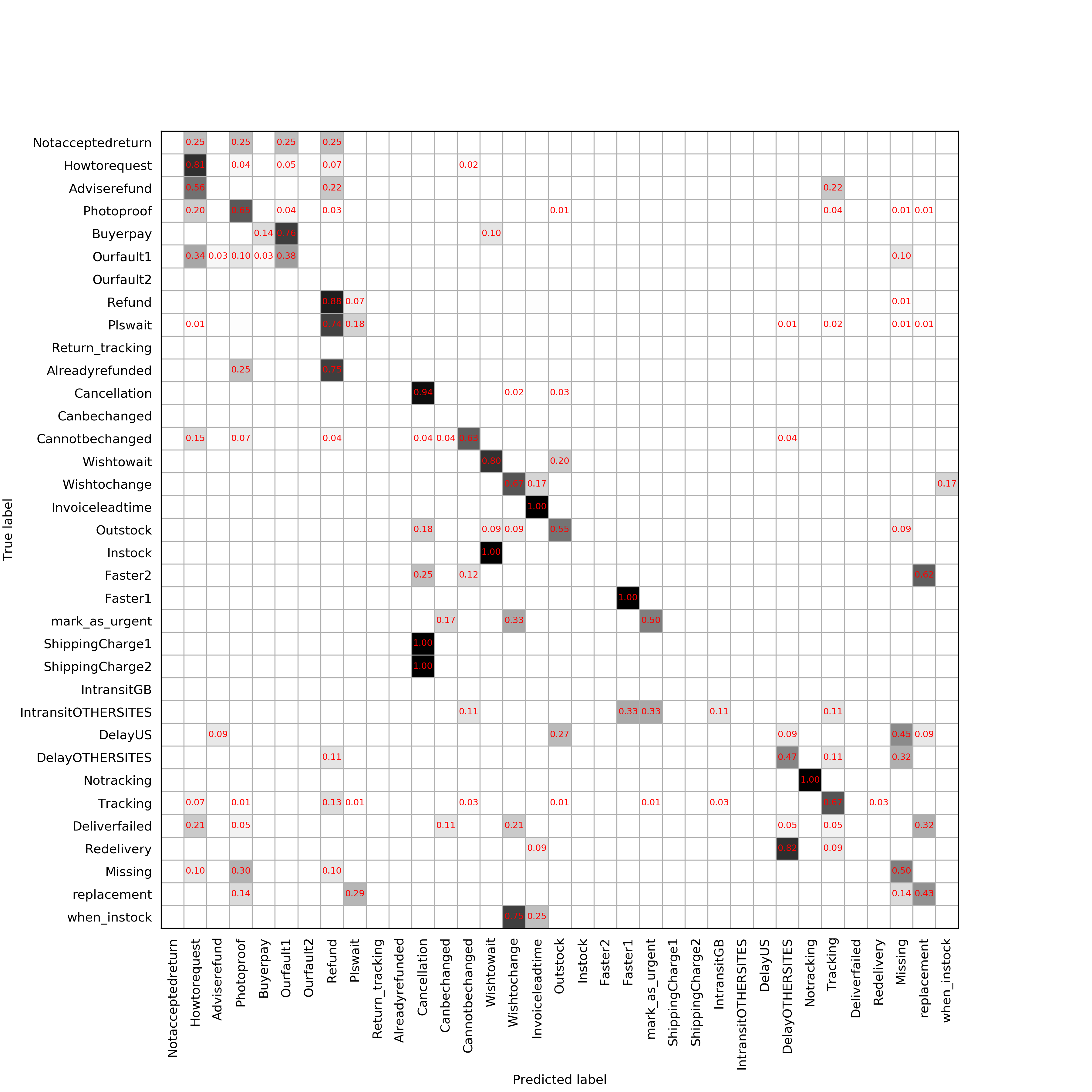}
\caption{Confusion matrix of Split Memory Network}
\label{fig:2}
\end{figure}

As shown in Table \ref{overall comparion results}, the gap between CA-Transformer and BERT demonstrates the importance of external knowledge. In other words, it is not enough to rely solely on powerful language model. Moreover, answers given by CAR-Transformer and CA-Transformer have overwhelming advantages over other methods in terms of the rate of accord with order conditions. This suggests that CAR-Transformer and CA-Transformer have better capacity in leveraging the order conditions information. CAR-Transformer utilizes conditions information in two components, conditions encoder and classifier. Although both CA-Transformer and TA-seq2seq are sequential language model, CA-Transformer performing better than TA-seq2seq suggests the superiority of the proposed conditions encoder. 
For further understanding the utilization of conditions, confusion matrices for classification models are displayed in Figures \ref{SA-Transformer}-\ref{fig:2}. For questions do not need conditions information to reply, such as questions whose answer are ``Howtorequest'' or ``Invoiceleadtime'', BERT performs well. While for most dialogues that need conditions information to reply, for example, dialogues which consult delivery, one can find that the models without knowledge aware have bad performance and incline to choose candidate answers with large amounts of samples, such as ``Missing'', ``Notracking'' and so on. Additionally, Split Memory Network tends to overfit the conditions information. It can be observed from Figure \ref{fig:2} that all ``Instock'' samples are classified to ``Wishtowait'' and 75\% ``when\_instock'' samples are classified to ``Wishtochange''. Since training samples whose answers are ``Instock'', ``when\_instock'', ``Wishtowait'' or ``Wishtochange'' have same condition value "out of stock", the conditions embedding may be very similar. CAR-Transformer usually can create a balance between buyers intent and conditions information. However, Figure \ref{SAR-Transformer} also indicates that CAR-Transformer is confused about the questions whose answers are ``Notacceptedreturn'', ``ShippingCharge1'', ``ShippingCharge2'' and so on. This is apparently due to these questions have few training samples. 

\subsubsection{The Effectiveness of Conditions Reviser}

\begin{table*}[!htbp]
\renewcommand{\arraystretch}{1.3}
\caption{Some representational cases of CAR-Transformer and CA-Transformer}
\label{cases}
\centering
\begin{tabular}{p{11em}p{5em}p{5em}p{5em}p{5em}p{6em}}
\toprule
\bfseries question & \bfseries defective conditions & \bfseries true conditions & \bfseries revised conditions & \bfseries prediction of CAR-Transformer & \bfseries prediction of CA-Transformer\\
\hline
Thank you for the update, I will wait for the new stock to come in & In stock & Out of stock & Out of stock & Wishtowait & Refund \\
\hline
Hi, could you please advise whether you have received my returned order? It's nearly two weeks since i posted it off to you. Thanks & Express missing, no return status & Express status is normal, not receive return goods & Express status is normal, not receive return goods & Plswait & Missing \\
\hline
Hi, I have 5 t-shirts on order and they haven't arrived as yet. Could you provide an update please. & Normal express, express status is redelivery & Expedite express, express status is redelivery & Expedite express, express status is redelivery & Redelivery & Notracking \\
\hline
Shirt arrived looks good, but too big as have another shirt similar. I need 17" collar, so XL would be better. Not opened package, so can I swap it for a smaller one? & Shipped: unknow, delivery status: unknown & Shipped: yes, delivery status: normal & Shipped: yes, delivery status: normal & Adviserefund & Plswait \\
\hline
Cancelled this order due to posting and packing churches when free delivery in UK was stated. I am disappointed that this order has been dispatched already & shipped & shipped & shipped & Cannotbechanged & Cancellation\\
\hline
My daughter needs the t-shirt for her show next Saturday 24th November, could you please dispatch asap, I'd be very grateful & shipped & Unshipped & Unshipped & mark as urgent & Notracking \\
\hline
I really need the high visible vest for the 6th of October this month ... if is possible it would it be great !! & shipped & shipped & shipped & Faster1 & Notracking\\
\bottomrule
\end{tabular}
\end{table*}

\begin{table*}
\centering
\caption{overall comparison results of conditions aware models on examples with defective conditions information}
\label{wrong_states_results}
\begin{tabular}{cccccccc}
	\toprule
    \multirow{2}{*}{Models} & \multicolumn{5}{c}{Automatic Evaluations} & \multicolumn{2}{c}{Human Evaluations} \\
	\cmidrule(r){2-6}  \cmidrule(r){7-8}
		& BLEU-1 & BLEU-2 & BLEU-3 & BLEU-4 & Accuracy & Rate of accord with conditions & Satisfaction \\
	\midrule
	SC-LSTM    & 36.57\% & 29.33\% & 26.56\% & 24.32\% & - & 38.85\%(0.17) & 1.20(0.30) \\
	TA-seq2seq & 55.89\% & 51.11\% & 48.34\% & 46.46\% & - & 58.83\%(0.12) & 3.39(0.17)\\
	Split Memory Network    & 41.77\% & 40.81\%& 40.10\% & 39.95\% & 39.86\% & 57.76\%(0.14) & 3.26(0.24) \\
	\midrule
	CA-Transformer & 58.30\% & 58.21\% & 57.52\% & 57.32\% & 57.25\% &60.37\%(0.12) & 3.55(0.08)\\
	CAR-Transformer & 64.93\% & 64.24\% & 64.02\% & 63.68\% & 63.60\% & 68.82\%(0.11) & 4.01(0.10)\\
	\bottomrule
\end{tabular}
\end{table*}

\begin{figure}[!htbp]
\centering
\includegraphics[width=3.5in]{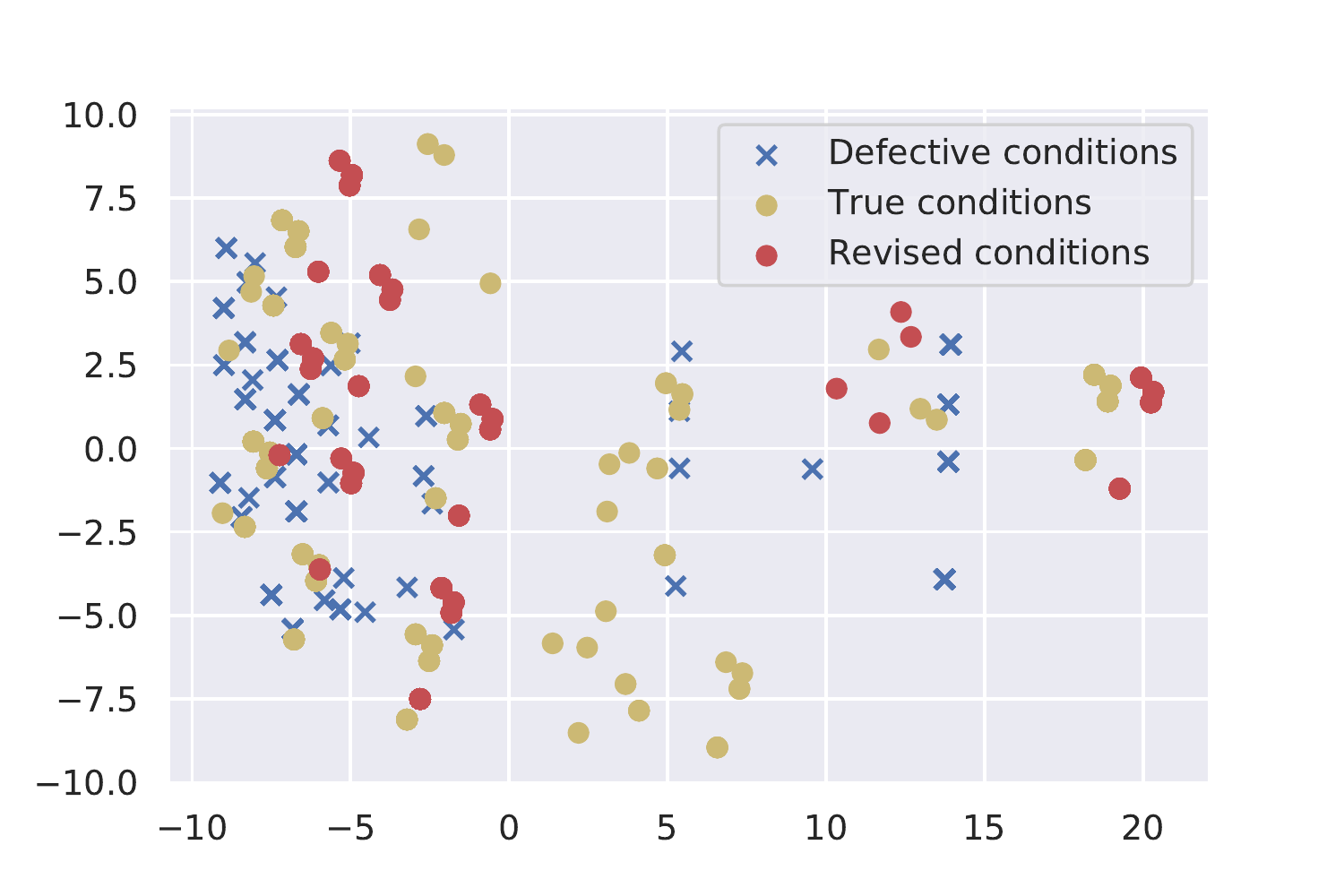}
\caption{Principal component analysis of conditions embedding vectors}
\label{pca}
\end{figure} 

As can be observed in Table \ref{overall comparion results}, CAR-Transformer has substantial gains over CA-Transformer. This suggests that the defective conditions information may affect the model's accuracy and the conditions reviser may reduce this impact. To provide some qualitative insights into the conditions reviser, some representational cases of CAR-Transformer and CA-Transformer are displayed in Table \ref{cases}. Note that the dialogue history and condition values that unrelated to reply question are omitted for the limitation of the space. Samples in rows 1 to 4 have wrong or unknow condition values and only CAR-Transformer predict correctly by revising condition values. Samples in rows 5 to 7 have right condition values, however, only CAR-Transformer predict correctly. Observed from Table \ref{cases} and Figures \ref{SA-Transformer}, CA-Transformer tends to weaken the function of conditions information. A possible explanation is that CA-Transformer pays less attention to conditions embedding because of the existence of defective condition values. However, CAR-Transformer can revise 71.13\% wrong or missing condition values in the test set to correct, and thus it can make good use of conditions information. Figure \ref{pca} shows the principal component analysis of three different types of conditions embedding vectors. It can be found that the revised conditions embeddings tend to the same distribution as correct conditions embeddings. The results of condition aware models on examples with defective conditions are also provided in Table \ref{wrong_states_results}. The so-obtained results indicate that defective condition values can be revised by the proposed conditions reviser to a certain degree.

\subsection{Supplementary experiment on personalized bAbI dialog dataset}
\begin{table}
\caption{Test results across models and tasks of personalized bAbI dialog dataset when masking user profiles}

\centering 
\label{personalizedt}
\begin{tabular}{p{3em}p{6em}p{4em}p{4em}p{6em}}
	\toprule
	Tasks	& Supervised Embeddings & Memory Network & CAR-Transformer \\
	\midrule
	PT1   & 17.34\% & 20.01\% & 80.66\%\\
	PT2 & 11.07\% & 18.76\% & 85.34\%\\
	PT3 & 9.04\% & 44.16\% & 44.84\%\\
	PT4 & 4.47\% & 43.23\% & 50.27\%\\
	PT5 & 10.36\% & 29.57\% & 76.19\%\\	
	\bottomrule
\end{tabular}
\end{table}

\begin{table}[!htbp]
\renewcommand{\arraystretch}{1.3}
\caption{user profile inferring results of personalized bAbI dialog dataset}
\label{personalizationinfer}
\centering
\begin{tabular}{ccc}
\toprule
\bfseries Task & \bfseries Best Accuracy & \bfseries Number of utterances used\\
\midrule
PT1 & 96.39\% & 2  \\

PT2 & 96.53\% & 2 \\

PT3 & 64.02\% & 10\\

PT4 & 95.66\% & 2\\

PT5 & 63.27\% & 14\\
\bottomrule
\end{tabular}
\end{table}

In this subsection, the proposed CAR-Transformer is extended to personalized bAbI dialog dataset \cite{joshi2017personalization}, which is a public multi-turn dialog corpus in a restaurant reservation scenario. It introduces additional four profiles (gender, age, dietary preference and favorite food) and the utterances are relevant to the user profiles. The bot is required to select an appropriate response from candidate set. Five separate tasks are introduced along with the dataset. Tasks 1 and 2 test the model's ability to indirectly track dialog state. Tasks 3 and 4 check whether the model can sort and use facts of restaurants. Task 5 tests the capabilities of all the above aspects of the model. Tasks 1, 2 and 4 only have gender and age information of users, task 3 and 5 have all attributes. More details of this dataset can be found in \cite{joshi2017personalization}. Instead of giving user profiles directly, we assume that all user profiles are unknown and use the proposed conditions reviser to infer the user profiles according to dialogue history. The main differences between our customer service dataset and the personalized bAbI dialog dataset lies in there is no correlation between profile attributes in personalized bAbI dialog dataset since the profile attributes are randomly sampled from a list of possible values. What's more, the user profiles "dietary preference" and "favorite food" in personalized bAbI dialog dataset are provided for bot to make choice between restaurants rather than associating with the content of dialogue directly.  

Based on above factors, the personalized bAbI dialog dataset is merely used for supplementary experiment. CAR-Transformer is compared with supervised embeddings and Memory Network \cite{weston2014memory} as in \cite{joshi2017personalization}. For CAR-Transformer, the max length of input utterance is set to 50 and only the last two turns of conversation history are considered. The hyperparameters of the models were selected on the validation sets.

Per-response accuracy (the percentage of responses in which the correct one is chosen out of all candidate ones) across all models are reported in Table \ref{personalizedt}. Table \ref{personalizationinfer} shows the best profiles prediction accuracy (percentage of correct inference for all user profiles) of conditions reviser on each task and how many utterances are used at least. The performance of CAR-Transformer is significant higher than other models, which indicates that the CAR-Transformer is able to infer and leverage user profiles to a certain extent. As can be observed in Table \ref{personalizationinfer},  for gender and age, they can be reasoned by the conditions reviser according to the style of language easily; but for dietary preference and favorite food, because of the diversity of choices (there are 2 types for dietary preference and 14 types for favorite food) and implicit correspondence between utterances and attributes, they are harder and need more conversation history to infer.

\section{Conclusion}

In this paper, CAR-Transformer is proposed to select appropriate answer that compatible with both user intent and the conditions of the question. Specifically, this paper considers more general and realistic situation where the condition values are wrong or incomplete. The proposed conditions reviser can revise the wrong or incomplete condition values without knowing which one is wrong beforehand. We perform extensive experimental evaluations of the proposed approach on the real world dataset and extend the CAR-Transformer to infer the user profiles in personalized bAbI dialog dataset. The experimental results show the effectiveness of the proposed CAR-Transformer. 
The explicit knowledge will be investigated to Incorporated to the learning model for more effective knowledge correction \cite{wu2015knowledge,wu2017knowledge} in the future work. 

\bibliographystyle{IEEEtran}
\bibliography{bibtex}

\end{document}